\pdfoutput=1

\documentclass[11pt]{article}

\usepackage{ACL2023}

\usepackage{times}
\usepackage{latexsym}
\usepackage{placeins}

\usepackage[T1]{fontenc}

\usepackage[utf8]{inputenc}

\usepackage{microtype}

\usepackage{inconsolata}

\usepackage{graphicx}
\usepackage{tikz}
\usepackage{amsmath}
\usepackage{float}

\title{The Role of Global and Local Context in Named Entity Recognition}

\author{Arthur Amalvy \\
  Laboratoire Informatique d'Avignon\\
  \texttt{arthur.amalvy@univ-avignon.fr} \\\And
  Vincent Labatut$^*$ \\
  Laboratoire Informatique d'Avignon\\
  \texttt{vincent.labatut@univ-avignon.fr} \\\AND
  Richard Dufour$^*$\\
  Laboratoire des Sciences du Numérique de Nantes\\
  \texttt{richard.dufour@univ-nantes.fr} \\
  }

\begin{document}
\maketitle
\def\thefootnote{*}\footnotetext{These authors contributed equally.}\def\thefootnote{\arabic{footnote}}

\begin{abstract}
    Pre-trained transformer-based models have recently shown great performance when applied to Named Entity Recognition (NER). As the complexity of their self-attention mechanism prevents them from processing long documents at once, these models are usually applied in a sequential fashion. Such an approach unfortunately only incorporates local context and prevents leveraging global document context in long documents such as novels, which might hinder performance. In this article, we explore the impact of global document context, and its relationships with local context. We find that correctly retrieving global document context has a greater impact on performance than only leveraging local context, prompting for further research on how to better retrieve that context. 
\end{abstract}

\section{Introduction}
Named Entity Recognition (NER) is a fundamental task in Natural Language Processing (NLP), and is often used as a building block for solving higher-level tasks. Recently, pre-trained transformer-based models such as BERT~\citep{devlin-2019-bert} or LUKE~\citep{yamada-2020-luke} showed great NER performance and have been able to push the state of the art further.

These models, however, have a relatively short range because of the quadratic complexity of self-attention in the number of input tokens: as an example, BERT~\citep{devlin-2019-bert} can only process spans of up to 512 tokens. For longer documents, texts are usually processed sequentially using a rolling window. Depending on the document, this local window may not always include all the context needed to perform inference, which may be present at the global document level. This leads to prediction errors~\citep{stanislawek-2019-ner_glass_ceiling}: In NER, this often occurs when the type of an entity cannot be inferred from the local context. For instance, in the following sentence from the fantasy novel \textit{Elantris}, one cannot decide if the entity \texttt{Elantris} is a person (\texttt{PER}) or a location (\texttt{LOC}) without prior knowledge: 
\begin{quote}
    \textit{``Raoden stood, and as he did, his eyes fell on Elantris again.''}
\end{quote}

In the novel, this prior knowledge comes from the fact that a human reader can recall previous mentions of \texttt{Elantris}, even at a very long range. A sequentially applied vanilla transformer-based model, however, might make an error without a \textit{neighboring} sentence clearly establishing the status of \texttt{Elantris} as a city.

While some works propose to retrieve external knowledge to disambiguate entities~\citep{zhang-2022-ner_retrieval, wang-2021-ner_context_cooperative_learning}, external resources are not always available. Furthermore, external retrieval might be more costly or less relevant than performing document-level context retrieval, provided the document contains the needed information, which depends on the type of document.

Therefore, we wish to explore the relevance of document-level context when performing NER. We place ourselves at the sentence level, and we distinguish and study two types of contexts:

\begin{itemize}
    \item \textit{local context}, consisting of surrounding sentences. This type of context can be used directly by vanilla transformer-based models, as their range lies beyond the simple sentence. Fully using surrounding context as in~\citet{devlin-2019-bert} is, however, computationally expensive.
    \item \textit{global context}, consisting of all sentences available at the document level. To enhance NER prediction at the sentence level, we retrieve a few of these sentences and provide them as context for the model. 
\end{itemize}

We seek to answer the following question: is local context sufficient when solving the NER task, or would the model obtain better performance by retrieving global document context? 

To answer this question, we conduct experiments on a literary NER dataset we improved from its original version~\citep{dekker-2019-evaluation_ner_social_networks_novels}.  We release the annotation process, data and code necessary to reproduce these experiments under a free license\footnote{\url{https://github.com/CompNet/conivel/tree/ACL2023}}.

\section{Related Work}
\label{sec:related_works}

\subsection{Sparse Transformers}
Since the range problem of vanilla transformer-based models is due to the quadratic complexity of self-attention in the number of input tokens, several works on \textit{sparse transformers} proposed alternative attention mechanisms in hope of reducing this complexity~\citep{zaheer-2020-big_bird,wang-2020-linformer,kitaev-2020-reformer,tay-2020-sinkhorn_attention,tay-2020-synthesizer,beltagy-2020-longformer,choromanski-2020-performer,katharopoulos-2020-linear_transformer,child-2019-sparse_transformer}. While reducing self-attention complexity improves the effective range of transformers, these models still have issues processing very long documents~\citep{tay-2020-long_range_arena}.

\subsection{Context retrieval}
Context retrieval in general has been widely leveraged for other NLP tasks, such as semantic parsing~\citep{guo-2019-retrieval_semantic_parsing}, question answering~\citep{ding-2020-cogltx}, event detection~\citep{pouran-2021-doc_level_event_detection}, or machine translation~\citep{xu-2020-retrieval_machine_translation}.

In NER, context retrieval has mainly been used in an external fashion, for example by leveraging names lists and gazetteers~\citep{seyler-2018-external_knowledge_ner, liu-2019-ner_gazetteers}, knowledge bases~\citep{luo-2015-joint_ner_disambiguation} or search engines~\cite{wang-2021-ner_context_cooperative_learning, zhang-2022-ner_retrieval}. Meanwhile, we are interested in document-level context retrieval, which is comparatively seldom explored. While \citet{luoma-2020-ner_context} study document-level context, their study is restricted to neighboring sentences, i.e. local context.

\section{Method and Experiments}
\label{sec:method}

\subsection{Retrieval Heuristics}
\label{sec:retrieval_heuristics}
We wish to understand the role of both \textit{local} and \textit{global} contexts for the NER task. We split all documents in our dataset (described in Section \ref{sec:dataset}) into sentences. We evaluate both local and global simple heuristics of sentence retrieval in terms of NER performance impact. We study the following \textit{local} heuristics:
\begin{itemize}
    \item \texttt{before}: Retrieves the closest $k$ sentences at the left of the input sentence.
    \item \texttt{after}: Same as \texttt{before}, but at the right of the input sentence.
    \item \texttt{surrounding}: Retrieves the closest $\frac{k}{2}$ sentences on both sides of the input sentence.
\end{itemize}

And the following \textit{global} heuristics:
\begin{itemize}
    \item \texttt{random}: Randomly retrieves a sentence from the whole document.
    \item \texttt{samenoun}: Randomly retrieves a sentence from the set of all sentences that have at least one common noun with the input sentence\footnote{If the set of sentences with a common noun is empty, the \texttt{samenoun} heuristic does not retrieve any sentence.}. Intuitively, this heuristic will return sentences that contain entities of the input sentence, allowing for possible disambiguation. We use the NLTK library~\citep{bird-2009-nltk} to identify nouns.
    \item \texttt{bm25}: Retrieves sentences that are similar to the input sentences according to BM25~\citep{robertson-1994-bm25}. Retrieving similar sentences has already been found to increase NER performance~\citep{zhang-2022-ner_retrieval, wang-2021-ner_context_cooperative_learning}.
\end{itemize}

It has to be noted that global heuristics can sometimes retrieve local context, as they are not restricted in which sentences they can retrieve at the document level. For all configurations, we concatenate the retrieved sentences to the input. During this concatenation step, we preserve the global order between sentences in the document.

\subsection{Oracles}
For each heuristic mentioned in Section \ref{sec:retrieval_heuristics}, we also experiment with an \textit{oracle} version. The oracle version retrieves 16 sentences from the document using the underlying retrieval heuristic, and retain only those that enhance the NER predictions the most. We measure this enhancement by counting the difference in numbers of NER BIO tags errors made with and without the context. In essence, the oracle setup simulates a perfect re-ranker model, and allows us to study the maximum performance of such an approach.

\subsection{Dataset}
\label{sec:dataset}
To evaluate our heuristics, we use a corrected and improved version of the literary dataset of~\citet{dekker-2019-evaluation_ner_social_networks_novels}. This dataset is comprised of the first chapter of 40 novels in English, which we consider long enough for our experiments.

\paragraph{Dataset corrections} The original dataset suffers mainly from annotation issues. To fix them, we design an annotation guide inspired by CoNLL-2003~\citep{tjong-2003-conll_2003_ner} and apply it consistently using a semi-automated process:
\begin{enumerate}
    \item We apply a set of simple rules to identify obvious errors\footnote{See Appendix~\ref{sec:dataset-corrections-rules} for details.} (for example, non capitalized entities annotated as \texttt{PER} are often false positives). Depending on the estimated performance of each rule, we manually reviewed its choices before application.
    \item We manually review each difference between the predictions of a BERT~\citep{devlin-2019-bert} model finetuned on a slightly modified version of the CoNLL-2003 dataset~\citep{tjong-2003-conll_2003_ner}\footnote{We modified the CoNLL-2003 dataset to include honorifics as part of \texttt{PER} entities to be consistent with our annotation guidelines.} and the existing annotations.
    \item We manually correct the remaining errors.
\end{enumerate}

\paragraph{Further annotations} The original dataset only consists of \texttt{PER} entities. We go further and annotate \texttt{LOC} and \texttt{ORG} entities. The final dataset contains 4476 \texttt{PER} entities, 886 \texttt{LOC} entities and 201 \texttt{ORG} entities.

\subsection{NER Training}
\label{sec:experiments}
For all experiments, we use a pretrained $\text{BERT}_{\text{BASE}}$~\citep{devlin-2019-bert} model, consisting in 110 million parameters, followed by a classification head at the token level to perform NER. We finetune BERT for 2 epochs with a learning rate of $2 \cdot 10^{-5}$ using the \texttt{huggingface transformers} library~\citep{wolf-2020-transformers}, starting from the \texttt{bert-base-cased} checkpoint.

\subsection{NER evaluation}
We perform cross-validation with 5 folds on our NER dataset. We evaluate NER performance using the default mode of the \texttt{seqeval}~\citep{nakayama-2018-seqeval} python library to ensure results can be reproduced.

\section{Results}
\label{sec:results}

\begin{figure*}[htb!]
    \centering
    \begin{minipage}[b]{.45\textwidth}
        \includegraphics[width=\linewidth]{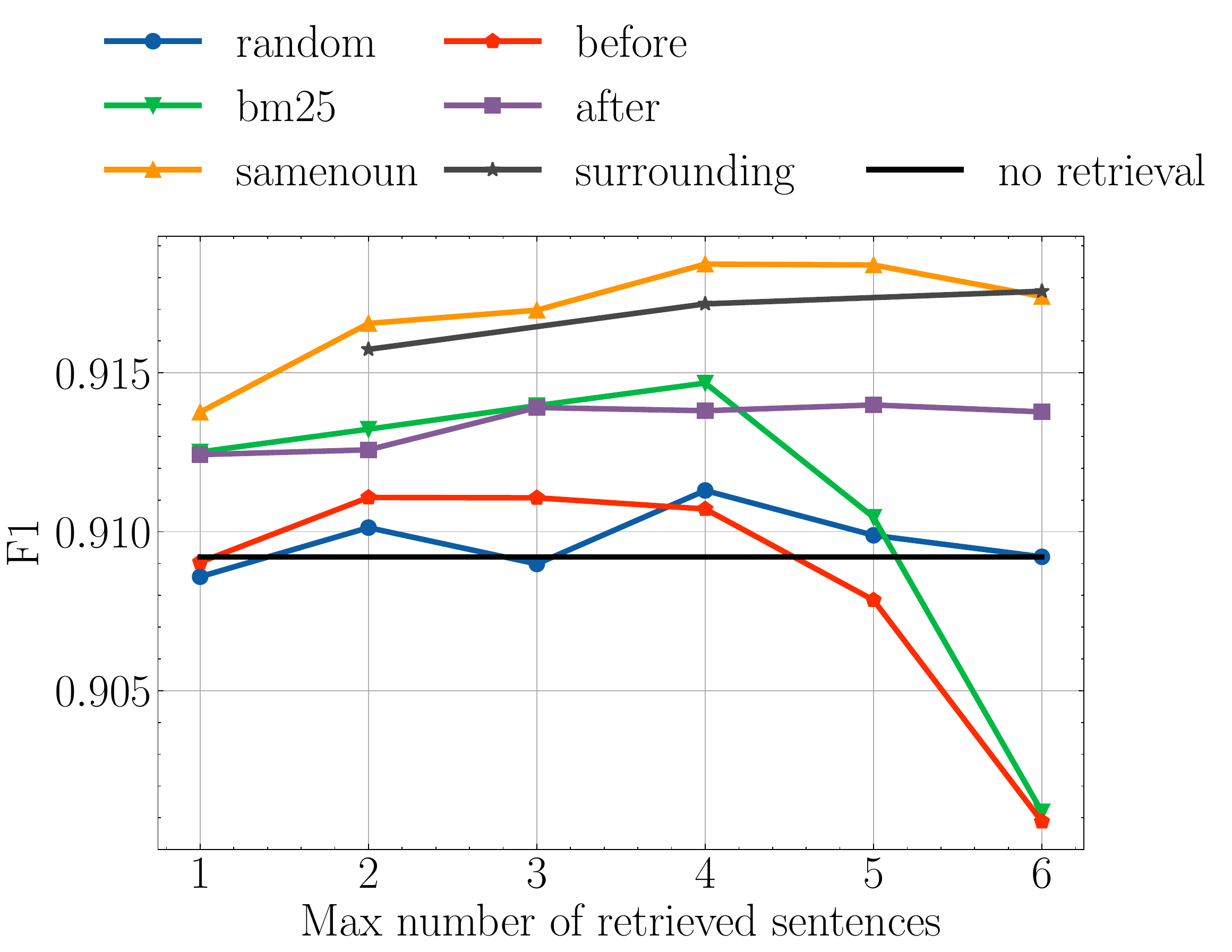}
        \caption{Mean F1 score versus max number of retrieved sentences for all retrieval heuristics across 3 runs.}
        \label{fig:main_results}
    \end{minipage}\qquad
    \begin{minipage}[b]{.45\textwidth}
        \includegraphics[width=\linewidth]{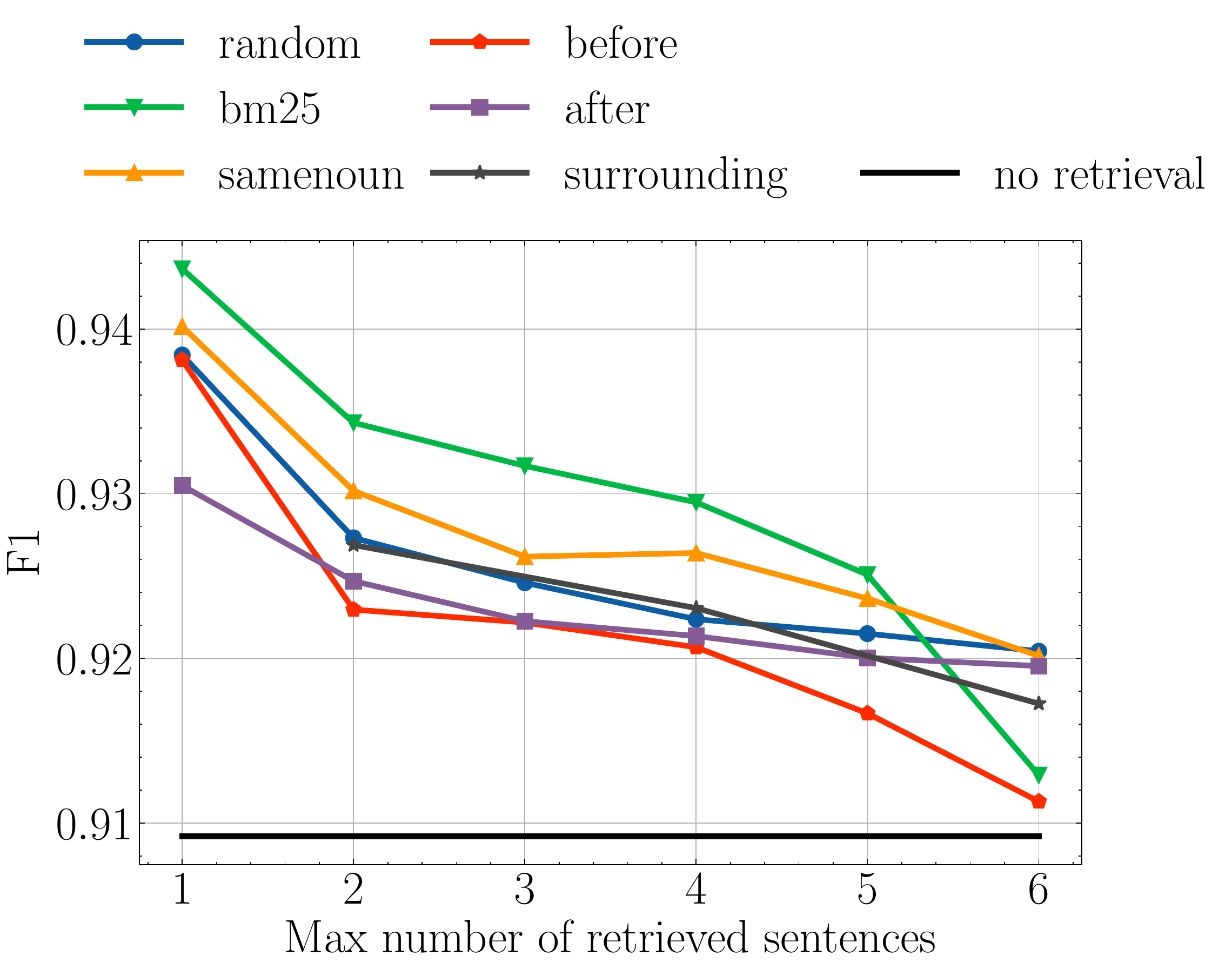}
        \caption{Mean F1 score versus max number of retrieved sentences across 3 runs for oracle versions of all retrieval heuristics.}
        \label{fig:oracle_results}
    \end{minipage}
\end{figure*}

\subsection{Retrieval heuristics}
The NER performance for retrieval heuristics can be seen in Figure~\ref{fig:main_results}.
The \texttt{samenoun} heuristic performs the best among global heuristics, whereas the \texttt{surrounding} heuristic is the best for local heuristics. While the top results obtained with both heuristics are quite similar, we consider global heuristics as naive retrieval baselines: they could be bested by more complex approaches, which might enhance performance even more.

Interestingly, the performance of both \texttt{before} and \texttt{bm25} heuristics decrease strongly after four sentences, and even drop behind the \texttt{no retrieval} baseline. For both heuristics, this might be due to retrieving irrelevant sentences after a while. The \texttt{bm25} heuristic is limited by the similar sentences present in the document: if there are not enough of them, the heuristic will retrieve unrelated ones. Meanwhile, the case of the \texttt{before} heuristic seems more puzzling, and could be indicative of a specific entity mention pattern that might warrant more investigations.

\subsection{Oracle versions}

\begin{figure*}[hbt!]
    \centering
    \includegraphics[width=\textwidth]{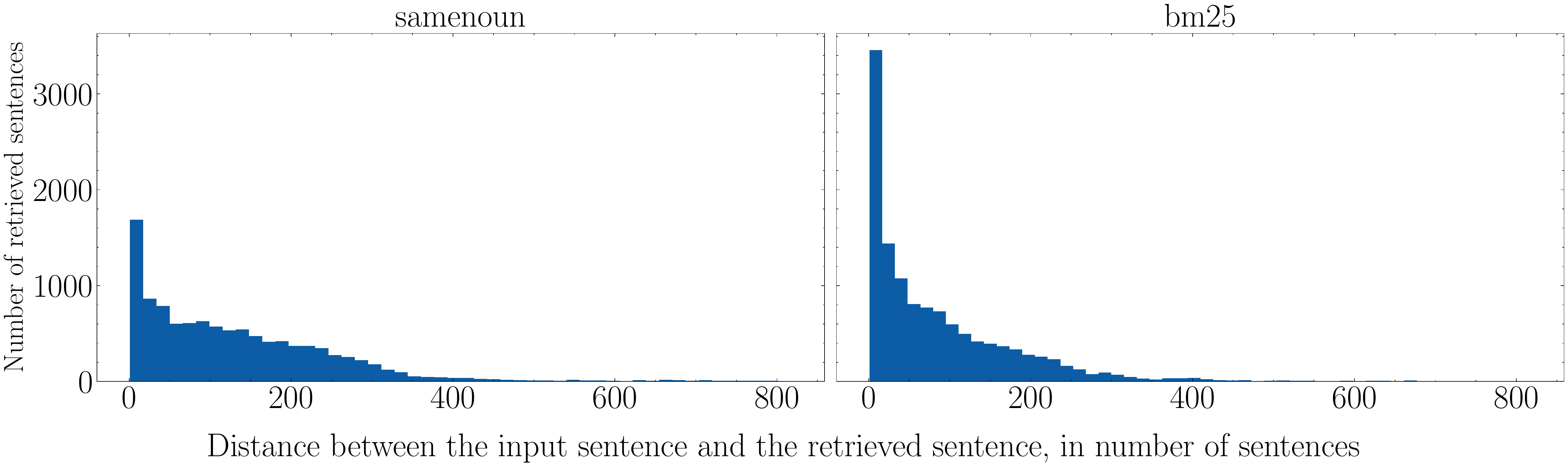}
    \caption{Distribution of the distance of retrieved sentences using the oracle versions of the \texttt{samenoun} and \texttt{bm25} heuristics. The \texttt{samenoun} heuristic retrieves fewer sentences overall, since it is possible for some sentence to not have a common noun with any other sentence of its document.}
    \label{fig:oracle_dists}
\end{figure*}

\begin{figure}[hbt!]
    \centering
    \includegraphics[width=\linewidth]{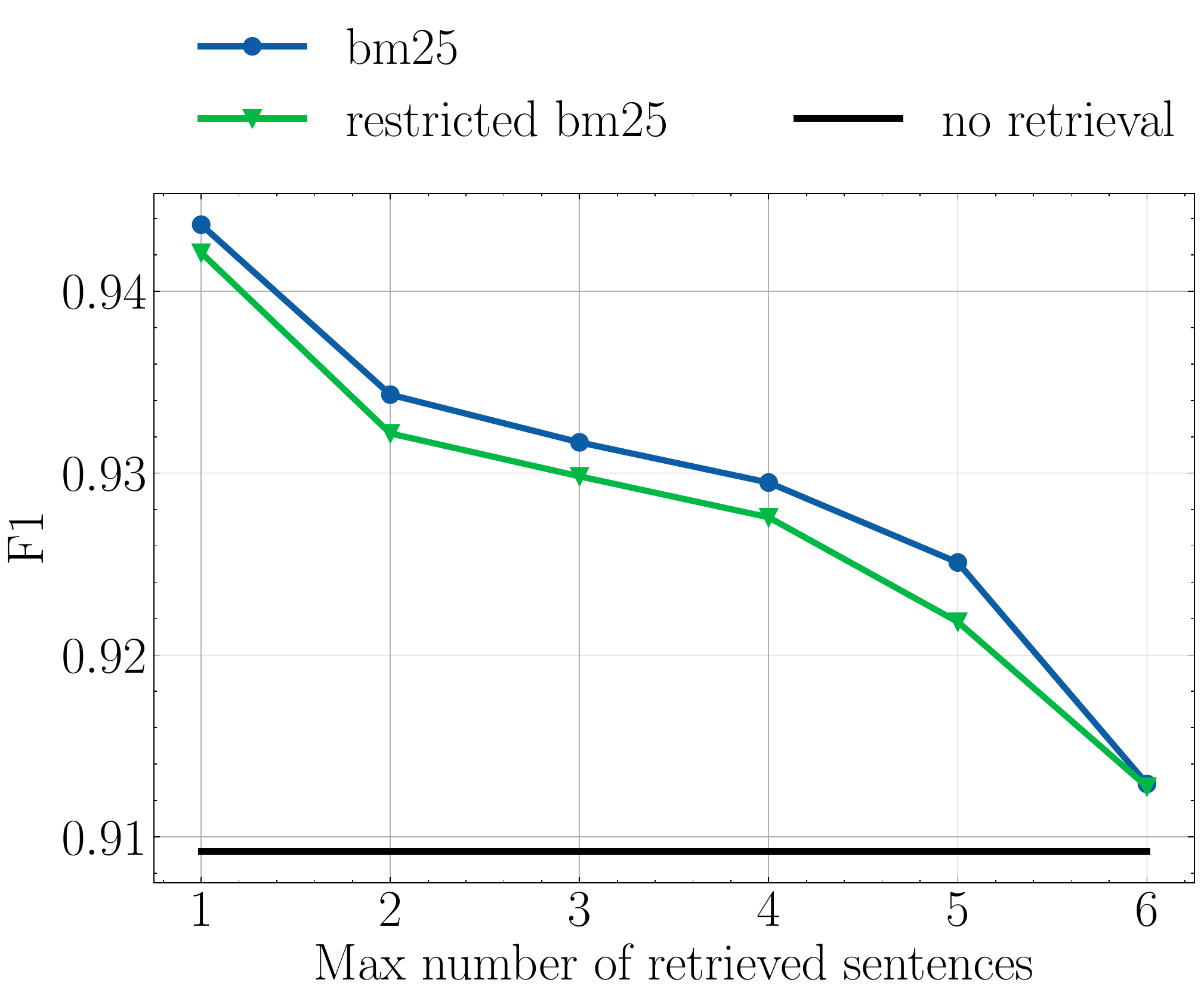}
    \caption{Mean F1 score versus number of retrieved sentences across 3 runs for the oracle version of the \texttt{bm25} heuristic, and the same heuristic restricted to distant context.}
    \label{fig:oracle_bm25_restricted}
\end{figure}

NER results with the oracle versions of retrieval heuristics can be found in Figure~\ref{fig:oracle_results}.

It is worth noting that the performance of the oracle versions of the heuristics always peaks when retrieving a single sentence. This might indicate that a single sentence is usually sufficient to resolve entity type ambiguities, but it might also be a result of the oracle ranking sentences individually, thereby not taking into account their possible combinations. 

Global heuristics perform better than local ones overall, with the oracle version of the \texttt{random} heuristic even performing better than both the \texttt{before} and \texttt{after} heuristics. These results tend to highlight the benefits of using global document context, provided it can be retrieved accurately.

\paragraph{Retrieved sentences} To better understand which sentences are useful for predictions when performing global retrieval, we plot in Figure~\ref{fig:oracle_dists} the distribution of the distance between sentences and their retrieved contexts for the oracle versions of heuristics \texttt{samenoun} and \texttt{bm25}. We find that 8\% and 16\% of retrieved sentences (for \texttt{samenoun} and \texttt{bm25}, respectively) are comprised within 6 sentences of their input sentence, while the other are further away, highlighting the need for long-range retrieval.

\paragraph{Local context importance} To see whether or not local context is an important component of NER performance, we perform an experiment where we restrict the oracle version of the \texttt{bm25} heuristic from retrieving local surrounding context. Results can be found in Figure~\ref{fig:oracle_bm25_restricted}. NER performance remains about the same without local context, which tends to show that local context is not strictly necessary for performance.

\section{Conclusion and Future Work}
In this article, we explored the role of local and global context in Named Entity Recognition. Our results tend to show that, for literary texts, retrieving global document context is more effective at enhancing NER performance than retrieving only local context, even when using relatively simple retrieval heuristics. We also showed that a re-ranker model using simple document-level retrieval heuristics could obtain significant NER performance improvements. Overall, our work prompts for further research in how to accurately retrieve global context for NER.

\section{Limitations}

We acknowledge the following limitations of our work:

\begin{itemize}
    \item While the oracle selects a sentence according to the benefits it provides when performing NER, it does not consider the interactions between selected sentences. This may lead to lowered performances when the several sentences are retrieved at once.
    \item The retrieval heuristics considered are naive on purpose, as the focus of this work is not performance. Stronger retrieval heuristics may achieve better results than presented in this article.
    \item The studied documents only consist in the first chapter of a set of novels. Using complete novel would increase the number of possible information to retrieve for the presented global heuristics.
\end{itemize}

\bibliographystyle{acl_natbib}
\bibliography{biblio}

\begin{thebibliography}{29}
\expandafter\ifx\csname natexlab\endcsname\relax\def\natexlab#1{#1}\fi

\bibitem[{Beltagy et~al.(2020)Beltagy, Peters, and
  Cohan}]{beltagy-2020-longformer}
I.~Beltagy, M.~E. Peters, and A.~Cohan. 2020.
\newblock \href {https://arxiv.org/abs/2004.05150} {Longformer: The
  long-document transformer}.
\newblock \emph{arXiv}, cs.CL:2004.05150.

\bibitem[{Bird et~al.(2009)Bird, Loper, and Klein}]{bird-2009-nltk}
S.~Bird, E.~Loper, and E.~Klein. 2009.
\newblock \emph{Natural Language Processing with Python}.
\newblock O'Reilly Media Inc.

\bibitem[{Child et~al.(2019)Child, Gray, Radford, and
  Sutskever}]{child-2019-sparse_transformer}
R.~Child, S.~Gray, A.~Radford, and I.~Sutskever. 2019.
\newblock \href {https://doi.org/10.48550/ARXIV.1904.10509} {Generating long
  sequences with sparse transformers}.
\newblock \emph{arXiv}, cs.LG:1904.10509.

\bibitem[{Choromanski et~al.(2020)Choromanski, Likhosherstov, Dohan, Song,
  Gane, Sarlos, Hawkins, Davis, Mohiuddin, Kaiser, Belanger, Colwell, and
  Weller}]{choromanski-2020-performer}
K.~Choromanski, V.~Likhosherstov, D.~Dohan, X.~Song, A.~Gane, T.~Sarlos,
  P.~Hawkins, J.~Davis, A.~Mohiuddin, L.~Kaiser, D.~Belanger, L.~Colwell, and
  A.~Weller. 2020.
\newblock \href {https://doi.org/10.48550/ARXIV.2009.14794} {Rethinking
  attention with performers}.
\newblock \emph{arXiv}, cs.LG:2009.14794.

\bibitem[{Dekker et~al.(2019)Dekker, Kuhn, and van
  Erp}]{dekker-2019-evaluation_ner_social_networks_novels}
N.~Dekker, T.~Kuhn, and M.~van Erp. 2019.
\newblock \href {https://doi.org/10.7717/peerj-cs.189} {Evaluating named entity
  recognition tools for extracting social networks from novels}.
\newblock \emph{PeerJ Computer Science}, 5:e189.

\bibitem[{Devlin et~al.(2019)Devlin, Chang, Lee, and
  Toutanova}]{devlin-2019-bert}
J.~Devlin, M.~Chang, K.~Lee, and K.~Toutanova. 2019.
\newblock \href {https://doi.org/10.18653/v1/N19-1423} {{BERT}: Pre-training of
  deep bidirectional transformers for language understanding}.
\newblock In \emph{Conference of the North {A}merican Chapter of the
  Association for Computational Linguistics: Human Language Technologies},
  volume~1, pages 4171--4186.

\bibitem[{Ding et~al.(2020)Ding, Zhou, Yang, and Tang}]{ding-2020-cogltx}
M.~Ding, C.~Zhou, H.~Yang, and J.~Tang. 2020.
\newblock \href
  {https://proceedings.neurips.cc/paper/2020/file/96671501524948bc3937b4b30d0e57b9-Paper.pdf}
  {{CogLTX}: Applying bert to long texts}.
\newblock In \emph{Advances in Neural Information Processing Systems},
  volume~33, pages 12792--12804.

\bibitem[{Guo et~al.(2019)Guo, Tang, Duan, Zhou, and
  Yin}]{guo-2019-retrieval_semantic_parsing}
D.~Guo, D.~Tang, N.~Duan, M.~Zhou, and J.~Yin. 2019.
\newblock \href {https://doi.org/10.18653/v1/P19-1082} {Coupling retrieval and
  meta-learning for context-dependent semantic parsing}.
\newblock In \emph{57th Annual Meeting of the Association for Computational
  Linguistics}, pages 855--866.

\bibitem[{Katharopoulos et~al.(2020)Katharopoulos, Vyas, Pappas, and
  Fleuret}]{katharopoulos-2020-linear_transformer}
A.~Katharopoulos, A.~Vyas, N.~Pappas, and François Fleuret. 2020.
\newblock Transformers are rnns: Fast autoregressive transformers with linear
  attention.
\newblock In \emph{Proceedings of the 37th International Conference on Machine
  Learning}, ICML'20.

\bibitem[{Kitaev et~al.(2020)Kitaev, Kaiser, and
  Levskaya}]{kitaev-2020-reformer}
N.~Kitaev, Ł. Kaiser, and A.~Levskaya. 2020.
\newblock \href {https://doi.org/10.48550/ARXIV.2001.04451} {Reformer: The
  efficient transformer}.
\newblock \emph{arXiv}, cs.LG:2001.04451.

\bibitem[{Liu et~al.(2019)Liu, Yao, and Lin}]{liu-2019-ner_gazetteers}
T.~Liu, J.~Yao, and C.~Lin. 2019.
\newblock \href {https://doi.org/10.18653/v1/P19-1524} {Towards improving
  neural named entity recognition with gazetteers}.
\newblock In \emph{57th Annual Meeting of the Association for Computational
  Linguistics}, pages 5301--5307.

\bibitem[{Luo et~al.(2015)Luo, Huang, Lin, and
  Nie}]{luo-2015-joint_ner_disambiguation}
G.~Luo, X.~Huang, C.~Lin, and Z.~Nie. 2015.
\newblock \href {https://doi.org/10.18653/v1/D15-1104} {Joint entity
  recognition and disambiguation}.
\newblock In \emph{2015 Conference on Empirical Methods in Natural Language
  Processing}, pages 879--888.

\bibitem[{Luoma and Pyysalo(2020)}]{luoma-2020-ner_context}
J.~Luoma and S.~Pyysalo. 2020.
\newblock \href {https://doi.org/10.18653/v1/2020.coling-main.78} {Exploring
  cross-sentence contexts for named entity recognition with {BERT}}.
\newblock In \emph{28th International Conference on Computational Linguistics},
  pages 904--914.

\bibitem[{Nakayama(2018)}]{nakayama-2018-seqeval}
H.~Nakayama. 2018.
\newblock \href {https://github.com/chakki-works/seqeval} {{seqeval}: A python
  framework for sequence labeling evaluation}.

\bibitem[{Pouran Ben~Veyseh et~al.(2021)Pouran Ben~Veyseh, Nguyen, Ngo~Trung,
  Min, and Nguyen}]{pouran-2021-doc_level_event_detection}
A.~Pouran Ben~Veyseh, M.~V. Nguyen, N.~Ngo~Trung, B.~Min, and T.~H. Nguyen.
  2021.
\newblock \href {https://doi.org/10.18653/v1/2021.emnlp-main.439} {Modeling
  document-level context for event detection via important context selection}.
\newblock In \emph{Conference on Empirical Methods in Natural Language
  Processing}, pages 5403--5413.

\bibitem[{Robertson(1994)}]{robertson-1994-bm25}
S.~E.~W. Robertson. 1994.
\newblock Some simple effective approximations to the 2-poisson model for
  probabilistic weighted retrieval.
\newblock In \emph{SIGIR '94}, pages 232--241.

\bibitem[{Seyler et~al.(2018)Seyler, Dembelova, Del~Corro, Hoffart, and
  Weikum}]{seyler-2018-external_knowledge_ner}
D.~Seyler, T.~Dembelova, L.~Del~Corro, J.~Hoffart, and G.~Weikum. 2018.
\newblock \href {https://doi.org/10.18653/v1/P18-2039} {A study of the
  importance of external knowledge in the named entity recognition task}.
\newblock In \emph{56th Annual Meeting of the Association for Computational
  Linguistics (Volume 2: Short Papers)}, pages 241--246.

\bibitem[{Stanislawek et~al.(2019)Stanislawek, Wróblewska, Wójcicka,
  Ziembicki, and Biecek}]{stanislawek-2019-ner_glass_ceiling}
T.~Stanislawek, A.~Wróblewska, A.~Wójcicka, D.~Ziembicki, and P.~Biecek.
  2019.
\newblock \href {https://doi.org/10.18653/v1/K19-1058} {Named entity
  recognition - is there a glass ceiling?}
\newblock In \emph{23rd Conference on Computational Natural Language Learning},
  pages 624--633.

\bibitem[{Tay et~al.(2020{\natexlab{a}})Tay, Bahri, Metzler, Juan, Zhao, and
  Zheng}]{tay-2020-synthesizer}
Y.~Tay, D.~Bahri, D.~Metzler, D.~Juan, Z.~Zhao, and C.~Zheng.
  2020{\natexlab{a}}.
\newblock \href {https://doi.org/10.48550/ARXIV.2005.00743} {Synthesizer:
  Rethinking self-attention in transformer models}.
\newblock \emph{arXiv}, cs.CL:2005.00743.

\bibitem[{Tay et~al.(2020{\natexlab{b}})Tay, Bahri, Yang, Metzler, and
  Juan}]{tay-2020-sinkhorn_attention}
Y.~Tay, D.~Bahri, L.~Yang, D.~Metzler, and D.~Juan. 2020{\natexlab{b}}.
\newblock \href {https://doi.org/10.48550/ARXIV.2002.11296} {Sparse sinkhorn
  attention}.
\newblock \emph{arXiv}, cs.LG:2002.11296.

\bibitem[{Tay et~al.(2020{\natexlab{c}})Tay, Dehghani, Abnar, Shen, Bahri,
  Pham, Rao, Yang, Ruder, and Metzler}]{tay-2020-long_range_arena}
Y.~Tay, M.~Dehghani, S.~Abnar, Y.~Shen, D.~Bahri, P.~Pham, J.~Rao, L.~Yang,
  S.~Ruder, and D.~Metzler. 2020{\natexlab{c}}.
\newblock \href {https://doi.org/10.48550/ARXIV.2011.04006} {Long range arena:
  A benchmark for efficient transformers}.
\newblock \emph{arXiv}, cs.LG:2011.04006.

\bibitem[{Tjong Kim~Sang and De~Meulder(2003)}]{tjong-2003-conll_2003_ner}
E.~F. Tjong Kim~Sang and F.~De~Meulder. 2003.
\newblock \href {https://aclanthology.org/W03-0419} {Introduction to the
  {C}o{NLL}-2003 shared task: Language-independent named entity recognition}.
\newblock In \emph{7th Conference on Natural Language Learning}, pages
  142--147.

\bibitem[{Wang et~al.(2020)Wang, Li, Khabsa, Fang, and
  Ma}]{wang-2020-linformer}
S.~Wang, B.~Z. Li, M.~Khabsa, H.~Fang, and H.~Ma. 2020.
\newblock \href {https://doi.org/10.48550/ARXIV.2006.04768} {Linformer:
  Self-attention with linear complexity}.
\newblock \emph{arXiv}, cs.LG:2006.04768.

\bibitem[{Wang et~al.(2021)Wang, Jiang, Bach, Wang, Huang, Huang, and
  Tu}]{wang-2021-ner_context_cooperative_learning}
X.~Wang, Y.~Jiang, N.~Bach, T.~Wang, Z.~Huang, F.~Huang, and K.~Tu. 2021.
\newblock \href {https://doi.org/10.18653/v1/2021.acl-long.142} {Improving
  named entity recognition by external context retrieving and cooperative
  learning}.
\newblock In \emph{59th Annual Meeting of the Association for Computational
  Linguistics and 11th International Joint Conference on Natural Language
  Processing}, volume~1, pages 1800--1812.

\bibitem[{Wolf et~al.(2020)Wolf, Debut, Sanh, Chaumond, Delangue, Moi, Cistac,
  Rault, Louf, Funtowicz, Davison, Shleifer, von Platen, Ma, Jernite, Plu, Xu,
  Le~Scao, Gugger, Drame, Lhoest, and Rush}]{wolf-2020-transformers}
T.~Wolf, L.~Debut, V.~Sanh, J.~Chaumond, C.~Delangue, A.~Moi, P.~Cistac,
  T.~Rault, R.~Louf, M.~Funtowicz, J.~Davison, S.~Shleifer, P.~von Platen,
  C.~Ma, Y.~Jernite, J.~Plu, C.~Xu, T.~Le~Scao, S.~Gugger, M.~Drame, Q.~Lhoest,
  and A.~M. Rush. 2020.
\newblock \href {https://www.aclweb.org/anthology/2020.emnlp-demos.6}
  {Transformers: State-of-the-art natural language processing}.
\newblock In \emph{Conference on Empirical Methods in Natural Language
  Processing: System Demonstrations}, pages 38--45.

\bibitem[{Xu et~al.(2020)Xu, Crego, and
  Senellart}]{xu-2020-retrieval_machine_translation}
J.~Xu, J.~Crego, and J.~Senellart. 2020.
\newblock \href {https://doi.org/10.18653/v1/2020.acl-main.144} {Boosting
  neural machine translation with similar translations}.
\newblock In \emph{Proceedings of the 58th Annual Meeting of the Association
  for Computational Linguistics}, pages 1580--1590.

\bibitem[{Yamada et~al.(2020)Yamada, Asai, Shindo, Takeda, and
  Matsumoto}]{yamada-2020-luke}
I.~Yamada, A.~Asai, H.~Shindo, H.~Takeda, and Y.~Matsumoto. 2020.
\newblock \href {https://doi.org/10.18653/v1/2020.emnlp-main.523} {{LUKE}: Deep
  contextualized entity representations with entity-aware self-attention}.
\newblock In \emph{Conference on Empirical Methods in Natural Language
  Processing}, pages 6442--6454.

\bibitem[{Zaheer et~al.(2020)Zaheer, Guruganesh, Dubey, Ainslie, Alberti,
  Ontanon, Pham, Ravula, Wang, Yang, and Ahmed}]{zaheer-2020-big_bird}
M.~Zaheer, G.~Guruganesh, K.~A. Dubey, J.~Ainslie, C.~Alberti, S.~Ontanon,
  P.~Pham, A.~Ravula, Q.~Wang, L.~Yang, and A.~Ahmed. 2020.
\newblock \href
  {https://proceedings.neurips.cc/paper/2020/file/c8512d142a2d849725f31a9a7a361ab9-Paper.pdf}
  {Big bird: Transformers for longer sequences}.
\newblock In \emph{Advances in Neural Information Processing Systems},
  volume~33, pages 17283--17297.

\bibitem[{Zhang et~al.(2022)Zhang, Jiang, Wang, Hu, Sun, Xie, and
  Zhang}]{zhang-2022-ner_retrieval}
X.~Zhang, Y.~Jiang, X.~Wang, X.~Hu, Y.~Sun, P.~Xie, and M.~Zhang. 2022.
\newblock \href {https://aclanthology.org/2022.coling-1.211} {Domain-specific
  {NER} via retrieving correlated samples}.
\newblock In \emph{Proceedings of the 29th International Conference on
  Computational Linguistics}, pages 2398--2404.

\end{thebibliography}

\appendix

\section{Dataset Details}

\subsection{Document Lengths}

\begin{figure}[h!]
  \centering
  \includegraphics[width=\linewidth]{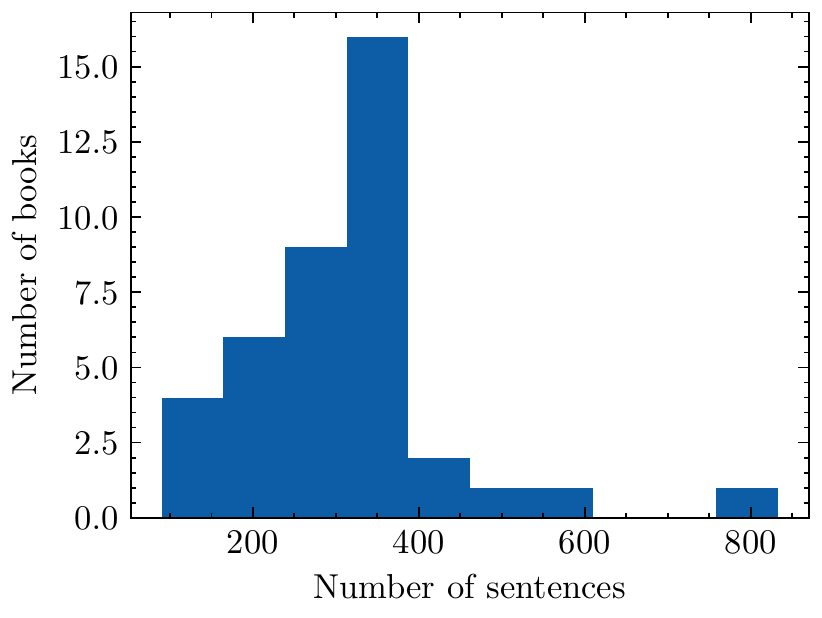}
  \caption{Distribution of the number of sentences in our enhanced version of the dataset from~\citet{dekker-2019-evaluation_ner_social_networks_novels}.}
  \label{fig:dekker_len}
\end{figure}

Our NER dataset is composed of documents longer that typical NER datasets such as CoNLL-2003~\citep{tjong-2003-conll_2003_ner}. Figure~\ref{fig:dekker_len} shows the distribution of the number of sentences of our NER dataset.

\subsection{Automatic Correction Rules}
\label{sec:dataset-corrections-rules}

We use the following rules to automatically identify obvious errors in the original dataset from~\citet{dekker-2019-evaluation_ner_social_networks_novels}. The original dataset only contained \texttt{PER} entities, so these rules only apply to them:

\begin{itemize}
    \item If a span appears in the list of characters from its novel but is not annotated as an entity, we investigate whether or not this is a false negative.
    \item Similarly, if a span annotated as an entity does not appear in the list of characters from its novel, we investigate whether or not it is a false positive.
    \item Finally, if a span is annotated as an entity but all of its tokens are not capitalized, we check if it is a false positive.
\end{itemize}

\section{Heuristics Results Breakdown by Precision/Recall}

\begin{figure*}[tb!]
    \centering
    \begin{minipage}[b]{.45\textwidth}
        \includegraphics[width=\linewidth]{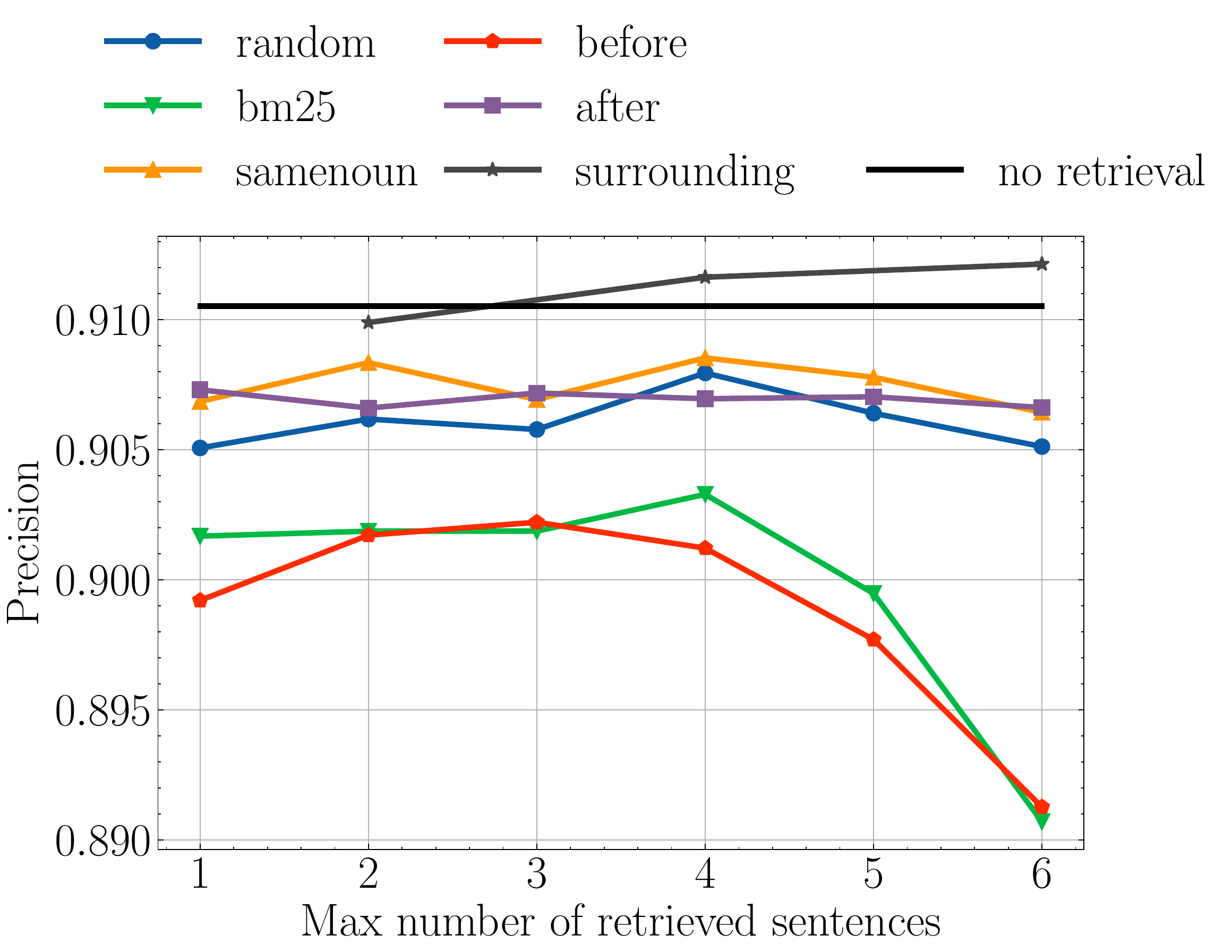}
        \caption{Mean precision versus max number of retrieved sentences for all retrieval heuristics across 3 runs.}
        \label{fig:main_precision_results}
    \end{minipage}\qquad
    \begin{minipage}[b]{.45\textwidth}
        \includegraphics[width=\linewidth]{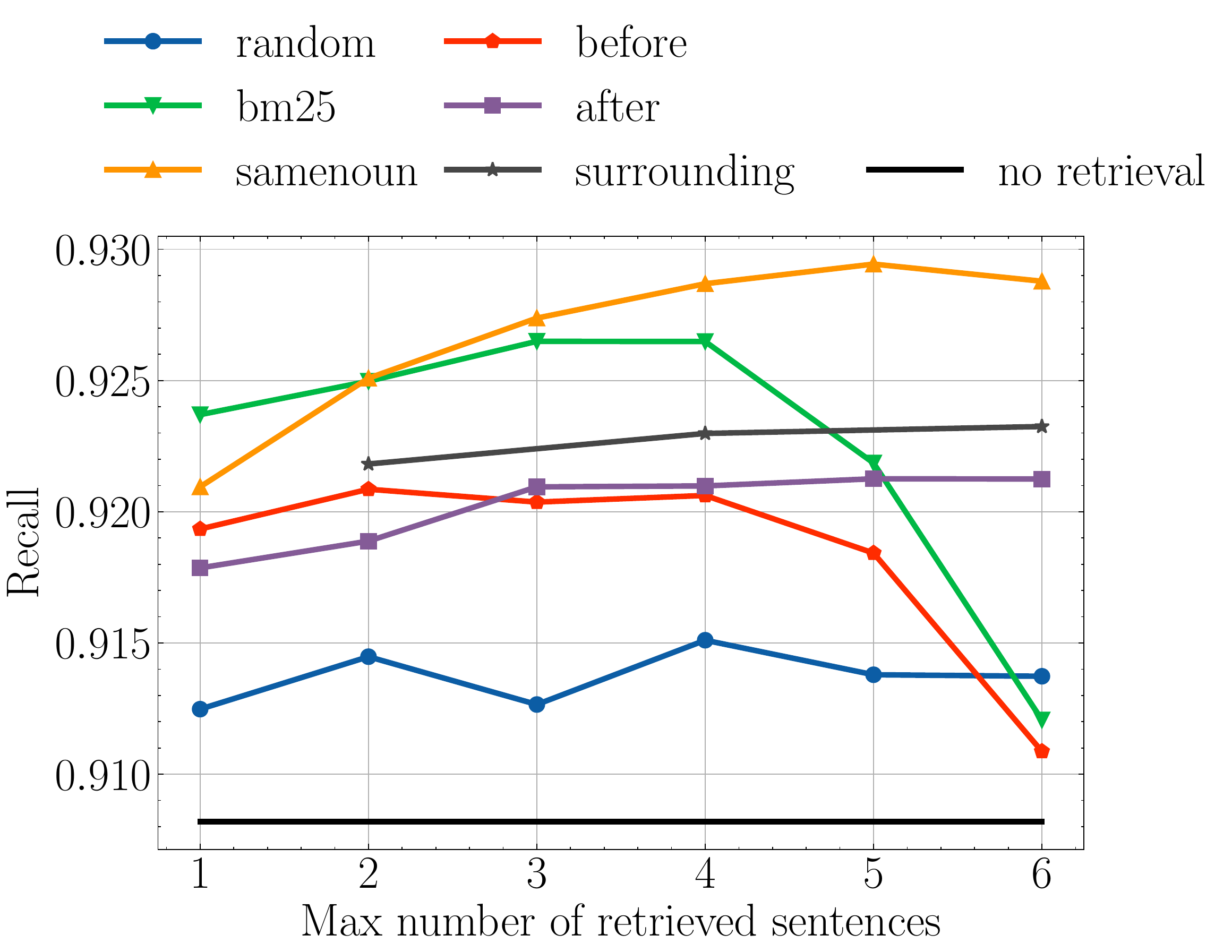}
        \caption{Mean recall versus max number of retrieved sentences for all retrieval heuristics across 3 runs.}
        \label{fig:main_recall_results}
    \end{minipage}  
\end{figure*}

Figures~\ref{fig:main_precision_results} and~\ref{fig:main_recall_results} show precision and recall for all retrieval heuristics. Interestingly, retrieval only has a positive effect on recall, with precision being lower than the baseline except for the \texttt{surrounding} heuristic.

\subsection{Oracle Versions}

\begin{figure*}[tb!]
    \centering
    \begin{minipage}[b]{.45\textwidth}
        \includegraphics[width=\linewidth]{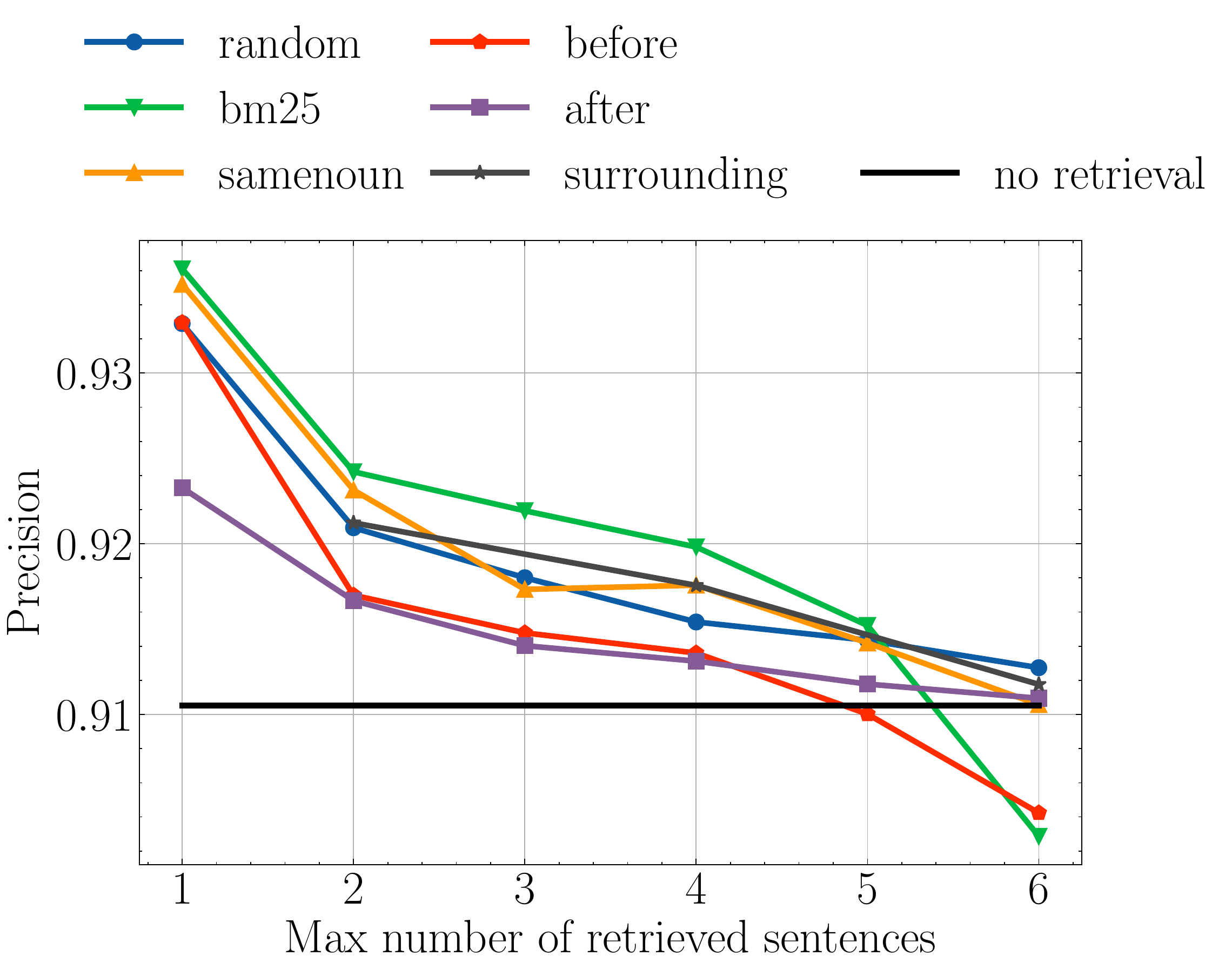}
        \caption{Mean precision versus max number of retrieved sentences across 3 runs for oracle versions of all retrieval heuristics.}
        \label{fig:oracle_precision_results}
    \end{minipage}\qquad
    \begin{minipage}[b]{.45\textwidth}
        \includegraphics[width=\linewidth]{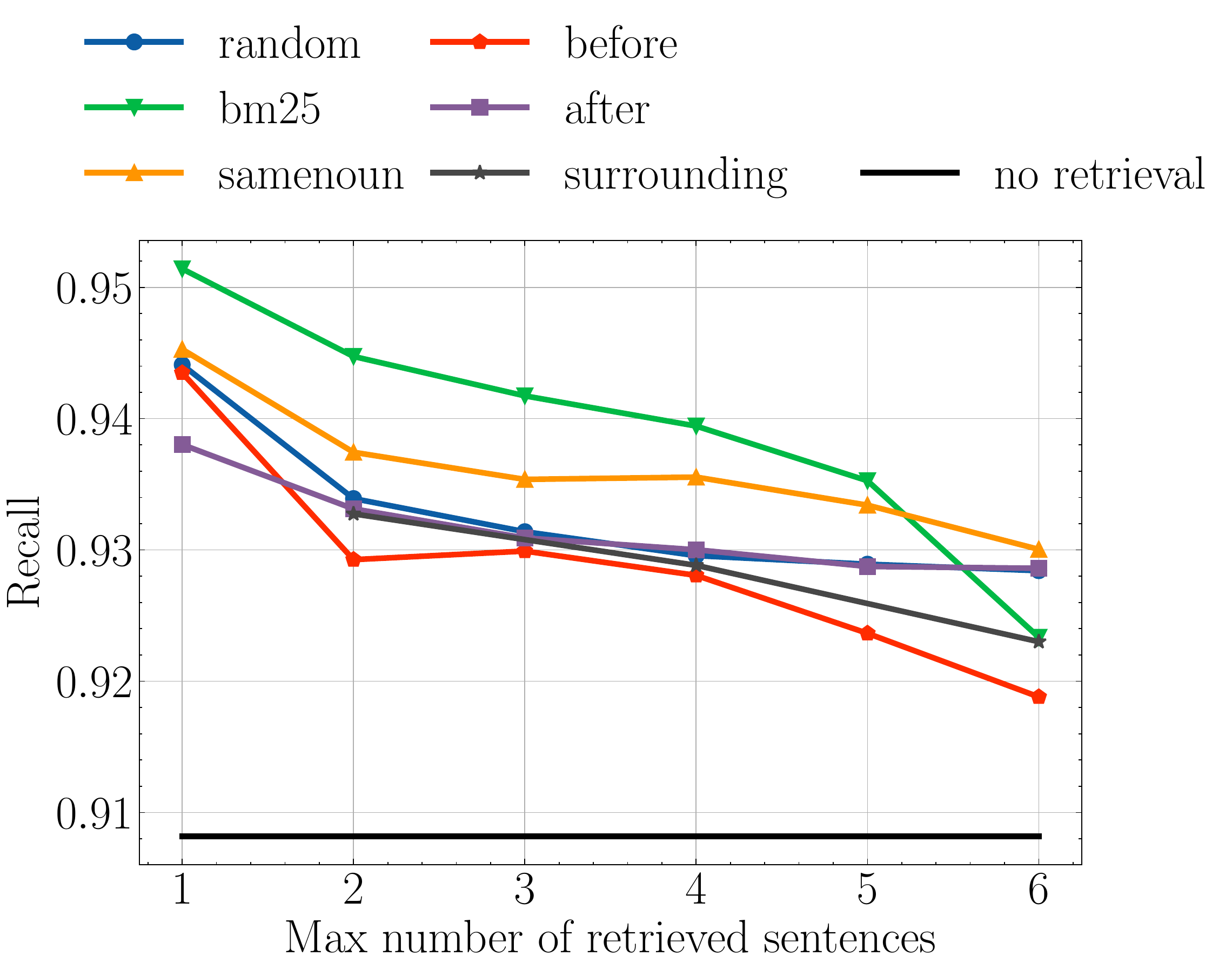}
        \caption{Mean recall versus max number of retrieved sentences across 3 runs for oracle versions of all retrieval heuristics.}
        \label{fig:oracle_recall_results}
    \end{minipage}  
\end{figure*}

Figures~\ref{fig:main_precision_results} and~\ref{fig:main_recall_results} show precision and recall for the oracle versions of all retrieval heuristics. While retrieval benefits recall more than precision, precision is still increased using retrieval. Together with the results from the regular heuristics, these results again highlight the potential performance gains of using a suitable re-ranker model to retrieve context.

\end{document}